\pdfoutput=1

\documentclass[]{article}

\usepackage{amsthm}
\usepackage{graphicx}
\usepackage{svg}
\usepackage[scaled=.98,sups,osf]{XCharter}
\usepackage[libertine,vvarbb,scaled=1.07]{newtxmath}
\usepackage{bm}
\usepackage{upref}
\usepackage{multirow}
\usepackage{multicol}
\usepackage[left=2.5cm,right=2.5cm,top=2.0cm,bottom=2.0cm]{geometry}
\usepackage{natbib}

\usepackage{cellspace}
\usepackage{amsmath}
\usepackage{mathtools}
\usepackage{physics} 
\usepackage[protrusion=true,tracking=true,kerning=true,expansion,final]{microtype}      
\usepackage{xcolor} 
\usepackage{mathtools} 
\usepackage{subcaption}
\usepackage{booktabs}       
\usepackage{multirow}
\usepackage{nicefrac}
\usepackage{enumitem}
\usepackage{varioref} 
\usepackage[pagebackref=true,breaklinks=true,colorlinks,bookmarks=true]{hyperref}
\usepackage[capitalise]{cleveref} 
\usepackage{soul}
\usepackage{diagbox}
\crefname{appsec}{appendix}{appendices}
\Crefname{appsec}{Appendix}{Appendices}

\definecolor{mydarkblue}{rgb}{0,0.08,0.45}
\definecolor{urlcolor}{rgb}{0,.145,.698}
\definecolor{linkcolor}{rgb}{.71,0.21,0.01}
\hypersetup{ %
	pdftitle={},
	pdfauthor={},
	pdfsubject={},
	pdfkeywords={},
	pdfborder=0 0 0,
	pdfpagemode=UseNone,
	breaklinks=true,  
	colorlinks=true,
	bookmarksnumbered=true,
	urlcolor=urlcolor,
	linkcolor=linkcolor,
	citecolor=mydarkblue,
	filecolor=mydarkblue,
	pdfview=FitH}

\renewcommand*{\backref}[1]{} 
\renewcommand*{\backrefalt}[4]{%
	\ifcase #1 %
	\or
	(cited on p. #2)%
	\else
	(cited on pp. #2)%
	\fi
}

\usepackage{textcomp}

\vbadness=\maxdimen
\usepackage{amsmath,amsfonts,bm}

\def\eqref#1{equation~\ref{#1}}

\def\1{\bm{1}}

\DeclareMathAlphabet{\mathsfit}{\encodingdefault}{\sfdefault}{m}{sl}
\SetMathAlphabet{\mathsfit}{bold}{\encodingdefault}{\sfdefault}{bx}{n}

\usepackage{hyperref}
\usepackage{url}
\usepackage{color}

\usepackage{graphicx}
\newcommand{\bb}[1]{\bm{\mathrm{#1}}}
\newtheorem{theorem}{Theorem}

\newtheorem{Example}
{Example}[theorem]

\usepackage{booktabs}  
\usepackage{tabularx}
\usepackage{setspace}
\setcitestyle{square}

\newcommand{\Phibold}{\mathbf{\Phi}}

\newcommand{\Ga}{\mathcal{G}_1}
\newcommand{\Gb}{\mathcal{G}_2}

\newcommand{\G}{\mathcal{G}}
\newcommand{\N}{\mathcal{N}}

\begin{document}
	
\title{Weisfeiler and Leman Go Infinite: Spectral and Combinatorial  Pre-Colorings}

\date{\vspace{-5.4ex}}

\author{Or Feldman$^{1}$, Amit Boyrski$^{1}$, Shai Feldman$^{1}$, Dani Kogan$^{1}$ , Avi Mendelson$^{1}$, Chaim Baskin$^{1}$\\
    {$^{1}$ Technion -- Israel Institute of Technology}\\
}

	\twocolumn[\maketitle]

	\begin{abstract}

	Graph isomorphism testing is usually approached via the comparison of graph invariants. Two popular alternatives that offer a good trade-off between expressive power and computational efficiency are combinatorial (i.e., obtained via the Weisfeiler-Leman (WL) test) and spectral invariants. While the exact power of the latter is still an open question, the former is regularly criticized for its limited power, when a standard configuration of uniform pre-coloring is used. This drawback hinders the applicability of Message Passing Graph Neural Networks (MPGNNs), whose expressive power is upper bounded by the WL test. Relaxing the assumption of uniform pre-coloring, we show that one can increase the expressive power of the WL test ad infinitum. Following that, we propose an efficient pre-coloring based on spectral features that provably increases the expressive power of the vanilla WL test.
	The above claims are accompanied by extensive synthetic and real data experiments. 
 	The code to reproduce our experiments is available at \url{https://github.com/TPFI22/Spectral-and-Combinatorial}.

		
	\end{abstract}


	\section{Introduction}

	Deep learning (DL) has become a method of choice for any machine learning task encountered in modern computer vision, natural language processing, and signal and image processing.  
    It has been particularly successful when dealing with Euclidean-structured data such as audio signals, images, and videos. 
    Attempts to generalize these approaches to non-Euclidean domains such as graphs and manifolds have led to the creation of a research area known as \textit{geometric deep learning} \cite{bronstein2017geometric}.
    This generalization of neural networks to non-Euclidean structured data was recently shown to work successfully in a wide-range of applications in computational social science \cite{monti2019fake, ying2018graph}, high-energy physics \cite{choma2018graph}, computational chemistry \cite{duvenaud2015convolutional}, 3D computer vision \cite{litany2017fmnet}, computational biology  \cite{ribeiro2017struc2vec} and medicine \cite{stokes2020anitbody}.
    
	\begin{figure}
		\begin{center}
			\includegraphics[width=\linewidth]{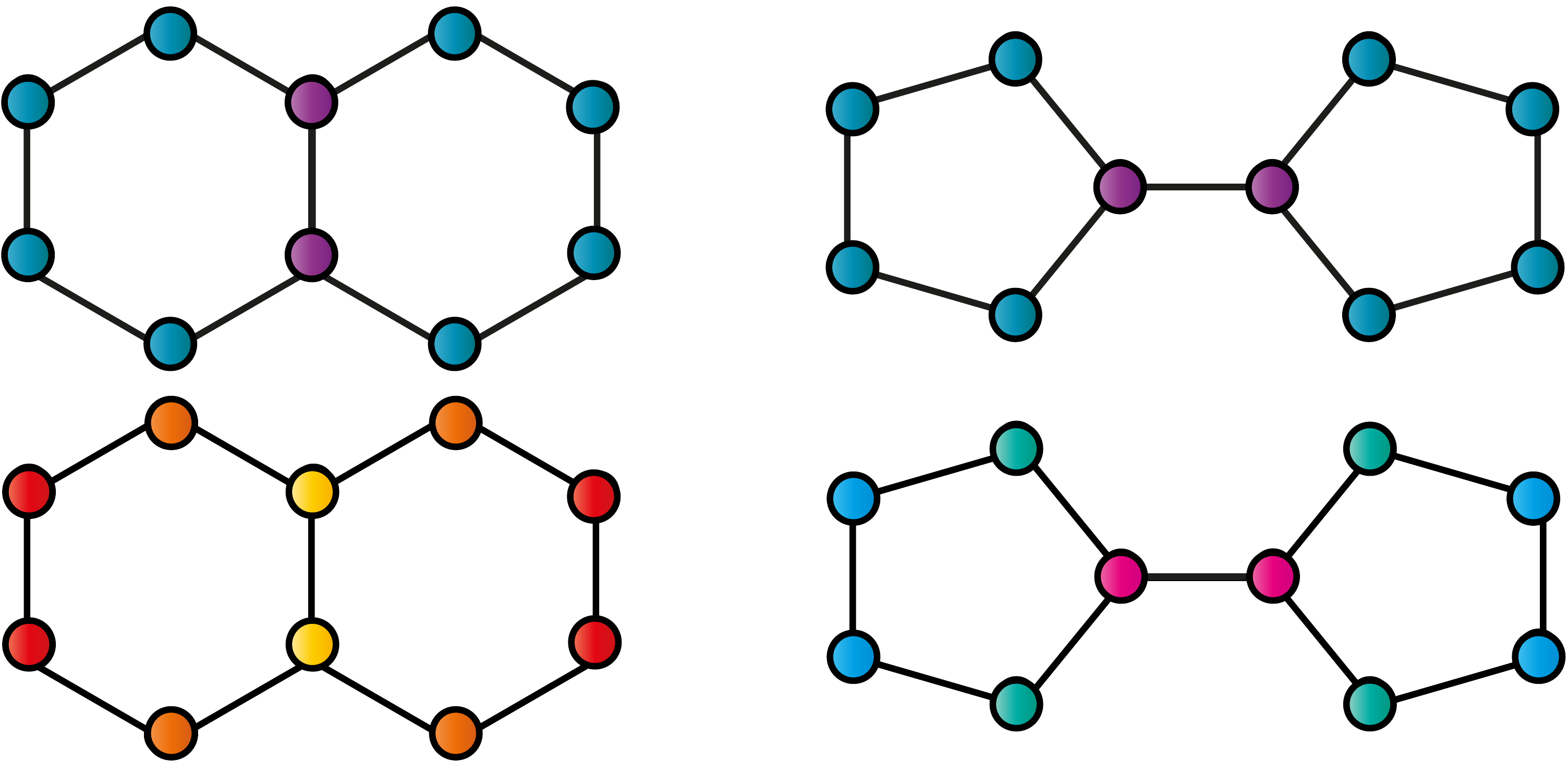}
		\end{center}
		\caption{The pair of graphs as colored by the degree coloring (upper) and the spectral coloring (lower).}
			\label{fig:intro}
	\end{figure}
    
    The term Graph Neural Networks (GNNs), as coined by \cite{bronstein2017geometric}, denotes neural networks designed to learn the non-Euclidean structure of graph data. The two main motivations that led to the modern GNN architectures \cite{zhou2020graph} are the notion of locality and weight sharing as used in CNNs  \cite{lecun1998gradient}, and graph representation learning \cite{hamilton2017representation}.
    
    Message Passing Neural Networks (MPNNs or MPGNNs, \cite{gilmer2017neural}) are collections of GNNs with common properties. MPGNNs use first order locality by recursively updating the features of each node from its neighborhoods' aggregated features. Then they create a descriptor for the graph by pooling all the node features together. MPGNNs are popular due to their efficiency \cite{balcilar2021breaking} and their ability to learn real world graph-structured data \cite{xu2018powerful}.
    
	Despite their success, MPGNNs are bounded in their expressive power (i.e., two different graphs may be encoded to the same descriptor by the same MPGNN). In fact, it is known that any two graphs that pass the WL test (described in detail in  \autoref{sec:preliminaries}) will be encoded by the same descriptor  \cite{xu2018powerful}. For example, MPGNNs cannot distinguish between the Decalin and Bicyclopentyl molecules graphs (\autoref{fig:1WLindistinguishable}) although their graphs are non-isomorphic  \cite{sato2020survey}.  
	Attempts have been made to improve the expressive power of MPGNNs by suggesting new and arguably complicated GNN architectures that are not bounded by the Weisfeiler-Leman (WL) test, e.g., by using high order networks, generalizing graphs to simplicial complexes, etc.
	
	We propose a new and general approach to improve the expressivity of MPGNNs. This approach is based on the traditional and relatively simple MPGNN architectures and does not require them to be changed at all. To that end, we suggest pre-coloring the nodes of a graph with an informative equivariant coloring, i.e., equivariant node features that are precomputed before the MPGNNs' learning process. We present a rigorous proof that this method can be used to improve the expressiveness of the WL test an infinite number of times. In addition, we present an instance of an equivariant coloring based on the spectral decomposition of the graph Laplacian that is also efficient to compute, explainable, and generates constant size features with respect to the graph size. \autoref{fig:intro} shows an example of the coloring of the Decalin and Bicyclopentyl molecules graphs with our suggested spectral pre-coloring, and the relatively simple degree pre-coloring. The example shows that the pair of graphs can be distinguished easily when using the spectral coloring compared to the degree coloring.   
	
    \paragraph{Contributions.}
    \begin{itemize}
	    \item We prove that the expressive power of WL can be improved ad infinitum by a sequence of equivariant pre-colorings and that each of the latter can be computed in polynomial time. Thus, the upper bound of the existing MPGNNs can be improved accordingly.
	    \item We suggest expressive and informative pre-coloring based on the spectral decomposition of the graph Laplacian, and explicitly prove that it improves the expressivity of the vanilla WL.
	    \item We perform extensive experiments showing that this simple extension improves the performance of various MPGNNs on different benchmarks.
	\end{itemize}

	\section{Preliminaries}\label{sec:preliminaries}
	
	\paragraph{Graph isomorphism.}
	An undirected graph of size $N$ is a pair $\G = (V,E)$ where $V=\{v_1...v_N\}$ is a set of vertices and $E$ is a set of edges. Each edge is a set of two vertices from $V$.
	We say that two graphs $\Ga=(V_1,E_1)$ and $\Gb=(V_2,E_2)$ are isomorphic if there exists bijection $\sigma : V_1 \rightarrow V_2$ s.t. $\{v_i,v_j\} \in E_1 \iff \{\sigma(v_i),\sigma(v_j)\} \in E_2$. There is no known polynomial time algorithm for determining isomoprhism between any arbitrary pair of graphs. Nevertheless, there are some classes of graphs (trees, planar) between which isomorphim can be determined using the polynomial time algorithm k-WL test \cite{kiefer2019weisfeiler,immerman1990describing}. Moreover, in \cite{grohe2012fixed} it was shown that for almost any class of graphs, the k-WL test is able to determine isomorphism. 
	\paragraph{Graph coloring.}
    Graph coloring is a mapping from a vertex and its graph to a label (color), from a known set of labels. We say that coloring $C$ \textit{refines} coloring $D$ if for any two graphs $\Ga$ and $\Gb$, and for any two vertices $v_1 \in V_1$, $v_2 \in V_2$ s.t. $C(v_1)=C(v_2)$, $D(v_1)=D(v_2)$. Ideally, we would like to find the following coloring: for each $v_1 \in V_1$,$v_2 \in V_2$, $C(v_1) = C(v_2) \iff$  there exists isomorphism  $\sigma : \Ga \rightarrow \Gb$ s.t., $\sigma (v_1) = v_2$.
    
	
	\paragraph{K-WL test.}
	The WL test of isomorphism is an algorithm for testing a necessary but insufficient condition for graph isomorphism.  Two graphs that do not pass the test are necessarily non-isomorphic. First, the algorithm assigns to each node the same color using the constant coloring $C^{0}_{WL}(v)=\text{CONST}$. Then the algorithm continues with iterations. At each iteration $i$, each node receives its neighbors’ colors and together with its own color, it generates a new color for the next iteration, i.e., $C^{i}_{WL}(v)=(C^{i-1}_{WL}(v),\{\{C^{i-1}_{WL}(x) | x \in \N(v)\}\})$, where `\{\{\}\}` denotes a multi-set, and $\N(v)$ denotes the set of neighbors of $v$. This process continues until convergence whereupon the colors are collected into a histogram. If the two graphs have different histograms, they failed the test and are called distinguishable. If after the convergence, the two histograms are the same, the graphs did not fail. Having thus passed the test, they are called indistinguishable.
	It was proved in \cite{bevilacqua2021equivariant} that $C^{i+1}_{WL}$ always \textit{refines} $C^i_{WL}$. The WL test can be extended to K-tuple coloring instead of vertex (1-tuple) coloring. This extension is called the K-WL test. It was proved in \cite{cai1992optimal} that any pair of graphs that are indistinguishable by k+1-WL are also indistinguishable by K-WL. Moreover, for any K$\geq$2, there exists a pair of graphs s.t. they are distinguishable by k+1-WL but indistinguishable by k-WL, i.e., k+1-WL is \textit{strictly more expressive} than k-WL, for K$\geq$2. The diagonal k-WL coloring on the graph vertices is defined to be $\Delta(k-WL)(v) = C_{k-WL}(v,...,v)$ where $C_{k-WL}$ is the coloring after the k-WL converges. It was proven in \cite{rattan2021weisfeiler} that $\Delta$(k+1-WL)  \textit{refines} $\Delta$(k-WL).

	
	\paragraph{Message Passing Graph Neural Networks.}
	MPGNNs are a specific type of GNNs. MPGNNs work in layers; each layer $l$ has its own Multi Layer Perceptron (MLP)$_l$ and iterates over the nodes in the graph. For each node, its neighbors' features are aggregated together with its own features using some aggregation operation.  The result of the aggregation is then used as input to the MLP of the current layer, and the output is the node's new features. To create a descriptor of the graph, the node features of each layer are aggregated separately, and the results are combined together. In other words, the node features of vertex $v$ after $l$ layers are
	$h_{v}^{(l)} = $ MLP$_{(l)}($UPDATE$(h_{v}^{(l-1)},$AGGREGATE$(\{\{h_{x}^{(l-1)} | x \in \N(v)\}\})))$
	where the graph descriptor is 
	$h_{\G} =$ COMBINE$(\{$AGGREGATE$(\{\{h_{x}^{(l)} | x \in \N(v)\}\}) | l \in \texttt{layers}\})$. It was proved in \cite{xu2018powerful} that the expressive power of MPGNNs is bounded by the expressive power of 1-WL, i.e., for any two graphs $\Ga$ and $\Gb$ s.t. $\Ga$ and $\Gb$ are indistinguishable by 1-WL, their descriptors created by any MPGNN will be equal. Moreover, it was proved that MPGNNs whose node features are aggregated using summation are strictly more expressive than MPGNNs that use other popular operations such as MAX and MEAN. The MPGNN based on summation is called Graph Isomorphism Network (GIN), and it has also been shown to produce SOTA results in addition to the theoretically superior expressiveness.
	
	\paragraph{Graph Laplacian}
	Let $\G = (V,E)$ be a weighted graph with an adjacency matrix denoted by $\bb{A}$. Given a function $\bb{x}\in \mathbb{R}^{|V|}$ on the vertices, the Dirichlet energy of the function $\bb{x}$ on the graph is defined to be
	\begin{equation}\label{eq:quadform}
	\bb{x}^\top \bb{L}\bb{x} = \sum_{(v,u)\in E}\bb{A}(v,u)\left(x(v)-x(u)\right)^2.
	\end{equation}
	
	The matrix $\bb{L}$ is the \textit{(combinatorial) graph Laplacian}, and is given by $\bb{L} = \bb{D}-\bb{A}$,
	where $\bb{D}$ is the \textit{degree matrix} i.e., diagonal matrix where $D(v,v)=|\N(v)|$. $\bb{L}$ is symmetric and positive semi-definite and, therefore, admits a spectral decomposition $\bb{L} = \Phibold\bb{\Lambda}\Phibold^\top$. Since the sum of each row in $\bb{L}$ is 0, $\lambda_1=0$ is always an eigenvalue of $\bb{L}$. The eigenpairs $(\bb{\phi}_i,\lambda_i)$ can be thought of as the graph analogues of `harmonic` and `frequency`. The graph Laplacian is the discrete generalization of the Laplace-Beltrami operator and hence it has similar properties to it.
	
	
	The spectrum of the graph Laplacian holds structural information about it. For example, the multiplicity of the zero eigenvalue represents the number of connected components in the graph. Another example is the second eigenvalue (counting multiple eigenvalues separately) that measures the connectivity of the graph \cite{spielman2009spectral}.
	We say that a pair of graphs are cospectral or cospectral with respect to the Laplacian, if their spectra of the Laplacians are equal.
	\paragraph{Heat kernel.}
	The heat kernel matrix describes the process of heat diffusion on the graph through time.
	The heat kernel at time $t$ for graph $\G=(V,E)$ is a $|V|\times|V|$ matrix where the element at the index $(u,v)$ is defined to be $H_{t}(u,v) = \Sigma^{|V|}_{i=1}e^{-\lambda_{i}t}\phi_i(u)\phi_i(v)$ where $\lambda_i$ is the i-th eigenvalue of the graph Laplacian and $\phi_i$ is its corresponding eigenvector.  $H_{t}(u,v)$ is the amount of heat transferred from node $u$ to node $v$ until time $t$. When the observed point in time $t$ tends to zero, the kernel is affected mostly by the local structures of the graphs. When the observed time point is relatively large, the global structure of the graphs becomes the dominant structure. 
	\section{Expressive power of 1-WL with pre-colorings}

	In \autoref{sec:preliminaries} we noted that the expressive power of 1-WL is limited. In particular, it is strictly limited by the expressive power of 3-WL. In this section we present a method to improve the expressive power of 1-WL using pre-coloring, i.e., coloring the graph before the iteration phase of 1-WL.
	If we pre-color 1-WL with coloring $C$, we mark the new algorithm as 1-$C$WL.
	
	\begin{theorem}
	\label{theorm:1}
		Let $R_1,R_2$ be two colorings s.t. $R_2$ \textit{refines} $R_1$ and $R_2$ is permutation equivariant. Accordingly, 1-$R_2$WL is at least as expressive as 1-$R_1$WL.
	\end{theorem}

    		Proof outline (the full proof can be found in \Cref{proof:t1}):
		\begin{enumerate}
		 \item We show that for any pair of isomorphic graphs, their histogram of 1-$R_2$WL is the same when using the permutation equivariant property of the coloring.
	
		\item We show that two graphs distinguishable by 1-$R_1$WL are also distinguishable by1-$R_2$WL. This is a corollary of the color refinement property of the 1-WL iterations.
		\end{enumerate}
		
		 For $R_1$ and $R_2$ that satisfy
        \autoref{theorm:1}, it is
        enough to find a single pair of graphs that are indistinguishable by 1-$R_1$WL but distinguishable by 1-$R_2$WL in order to prove strictness in expressive power.

        \begin{theorem}
        \label{theorm:2}
        Let $\Ga$,$\Gb$ be any two graphs. Their $\Delta$(k-WL) histograms are equal $\iff$ their $C_{k-WL}$ histograms are equal.
        \end{theorem}
        
        Proof outline (the full proof can be found in \Cref{proof:l1}):
		\begin{enumerate}
		 \item We prove the first direction by running k-WL for extra k-1 iterations after it converged. Following the structure of the k-WL, we show that the coloring of the graph is effectively "folded" onto the diagonal. Since these iterations happen after convergence, they do not change the final coloring, which means that the full graph histogram is encoded onto the diagonal tuples.
	
		\item We prove the second direction by the fact that in the initialization of k-WL, if a tuple is colored with the same color as a diagonal tuple, then it is necessarily a diagonal tuple. It, therefore, retains this property throughout the iterations.
		
		\end{enumerate}
        
	    \begin{theorem}
	    \label{theorm:3}
	        For any $K\geq2$, 1-$\Delta$(k+1-WL)WL is strictly more expressive than 1 -$\Delta$(k-WL)WL.
    	\end{theorem}
    	
    	The meaning of this theorem is that the expressive power of MPGNNs, which is provingly bounded by the expressive power of 1-WL, can be improved ad infinitium in the WL hierarchy using the right permutation equivariant pre-coloring as a pre-process before the MPGNN learning phase. According to \autoref{theorm:3}, the coloring can be obtained via the computation of $\Delta$(k-WL).
    	
    	In \autoref{sec:spectral} we give another example of such pre-coloring based on spectral features.
	
		Not every permutation equivariant coloring $C$ makes 1-$C$WL strictly more expressive than 1-WL.
		
		\begin{Example}
		If $D(u)= |\N(u)|$, i.e., the degree coloring, then 1-$D$WL is equal to 1-WL in terms of expressive power.
		\end{Example}

	\section{Spectral pre-coloring}
	\label{sec:spectral}
	
	\paragraph{Spectral WL.}
		We propose an expressive pre-coloring based on the graph spectrum, which can be used to color the nodes instead of the constant coloring of the 1-WL algorithm.
	    We will call this variant the \textit{spectral WL algorithm}. 
	    To calculate the pre-coloring, we first compute $m$ heat kernel matrices for evenly spaced points in time on the logarithmic scale. Then for each node $u$, we give the following color: $(H_{t_1}(u,u),...,H_{t_m}(u,u))$. Finally, we choose a constant amount of quantiles $r$  from the row of $u$ (ignoring the element on the diagonal) and append them in ascending order, e.g., $(({q^{t_1}_{1_u}}...{q^{t_1}_{r_u}}),...({q^{t_m}_{1_u}}...{q^{t_m}_{r_u}}))$, to the existing color of the node, to create the final coloring. In the example of the spectral coloring in \autoref{fig:intro}, nodes that have the same color have the same spectral features with $m=1$, $t=1$ and no quantiles. This simple setting is sufficient in order to compute the ideal equivariant coloring of the graphs.
	    \begin{theorem}
	    \label{theorm:4}
	        Spectral WL is strictly more expressive than 1-WL.
	    \end{theorem}
	
	Proof outline (the full proof can be found in \Cref{proof:t3}):
		\begin{enumerate}
		 \item We prove that \textit{spectral WL} is as expressive as 1-WL using \autoref{theorm:1} and the fact that any coloring \textit{refines} the constant coloring.
	
		\item We show concrete example where \textit{spectral WL} can distinguish between a pair of non-isomorphic graphs and 1-WL does not.
		\end{enumerate}
	
	\paragraph{Spectral features for GNNs.}
	This pre-coloring can be used to create initial node features for MPGNNs as a pre-process before the learning phase. Instead of applying the coloring we can append it to the existing node features of any graph. As hinted by \autoref{theorm:4}, in \autoref{sec:experiments} we will see that it is enough to add a relatively small feature vector, e.g., with 10 entries, to achieve great expressivity even for real world graphs with hundreds and thousands of nodes. One can, however, \textit{refine} the pre-processing by adding more quantiles and time samples.
	The features that we added to each node have the desirable property of being explainable, and they have the following meaning: For node $u$, the feature at entry $i\leq m$ is the amount of heat left at $u$ at time $t_i$ from the beginning of a diffusion process where all the nodes had $0$ heat and $u$ had exactly $1$. The features at entries $i>m$ represent the distribution of the heat diffusion through time on the other nodes.
	\paragraph{Scaling for large graphs.} For enormous graphs, calculating all the eigenvalues and eigenvectors is impractical. For large and small values of $t$, one can approximate $\bb{H}_t$ via spectral or spatial techniques. For large values of $t$ and $\lambda$, $e^{-t\lambda}$ become negligible, hence it is enough to rely on the $k$ smallest eigenvalues. For small values of $t$, $\bb{H}_t$ can be computed iteratively using explicit/implicit Euler iterations. Alternatively, we can use the model order reduction (MOR) technique to obtain approximate dynamics. The technique has been used successfully for computing isometry invariant descriptors for shape analysis \cite{bahr2018fast}. To that end, given the discretized heat equation
	\begin{equation}
	    \dot{\bb{x}} +\bb{L} \bb{x}=\bb{0},
	\end{equation}
	we use the eigendecomposition $\bb{L} = \bb{\Phi}\bb{\Lambda}\bb{\Phi}^\top$ to obtain 
	\begin{equation}
	    \bb{\Phi}^\top\dot{\bb{x}} +\bb{\Lambda}\bb{\Phi}^\top \bb{x}=\bb{0}.
	\end{equation}
	Since the dynamics are governed by the smaller eigenvalues of $\bb{L}$, we can truncate the eigendecomposition to obtain a lower-dimensional approximation of the dynamics. Denoting $\bb{w}_k = \bb{\Phi}_k^\top \bb{x}$, with $\bb{\Phi}_k$  being the first (smallest) $k$ eigenvectors of $\bb{L}$, we compute the approximated heat kernel at time $t$ by integrating 
	\begin{equation}\label{eq:approximateDynamics}
	    \dot{\bb{w}}_k +\bb{\Lambda} \bb{w}_k=\bb{0}
	\end{equation}
    up until the relevant time. \autoref{eq:approximateDynamics} is a simple PDE with a diagonal matrix $\bb{\Lambda}$ that can easily be solved via the unconditionally stable implicit Euler method.
    
    Despite the fact that we thus obtain only an approximation of the heat kernel, it is important to remember that our task is not to solve the heat equation but rather to provide useful spectral features for graph learning tasks. 
    In the related problem of non-rigid shape retrieval, descriptors obtained via the approximated dynamics (\autoref{eq:approximateDynamics}) actually provide better retrieval results compared to descriptors based on the full dynamics \cite{bahr2018fast}.
	\section{Experimental study on synthetic benchmarks}
	\label{sec:experiments}
\begin{figure}
		\begin{center}
			\includegraphics[width=\linewidth]{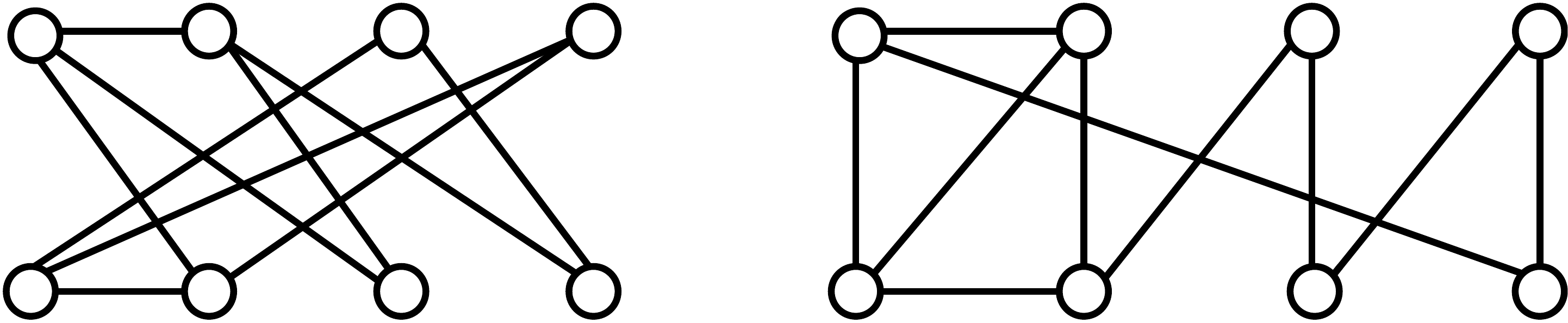}
		\end{center}
		\caption{Second pair of the original graphs. 1-WL distinguishable but cospectral with respect to the Laplacian.}
			\label{fig:cospectral}
	\end{figure}

	To demonstrate the improvement in expressivity that the spectral features add, we built two benchmarks, each of which is based on a single pair of graphs. The first pair of graphs is the Decalin and Bicyclopentyl molecule graphs that have the same 1-WL histogram \cite{sato2020survey}, but their spectrum is different. The second pair of graphs is shown in \autoref{fig:cospectral}, and they are distinguishable by 1-WL but cospectral with respect to the Laplacian. For each benchmark, we created 1000 examples by adding or removing a single edge at random from the original graphs and reordering their node indices randomly. For each benchmark, we split all the instances into training and test sets with ratio at a 9:1. The goal of a classifier for the benchmark is: Given a graph from the test set, identify the original graph from which it was perturbed. We trained GIN \cite{xu2018powerful}, GCN \cite{kipf2016semi}, GraphSAGE \cite{hamilton2017inductive} and GAT \cite{velivckovic2017graph} and their appropriate Spectral Pre-processed (SP) classifiers with the same settings of five message passing layers, a hidden dimension of 64, a learning rate of 0.01  and spectral features from 10 points in time using only the maximum quantile, for 100 epochs.
	We repeated the experiment 100 times and report the average accuracy and standard deviation
of each classifier.
	\begin{figure}
		\begin{center}
			\includegraphics[width=\linewidth]{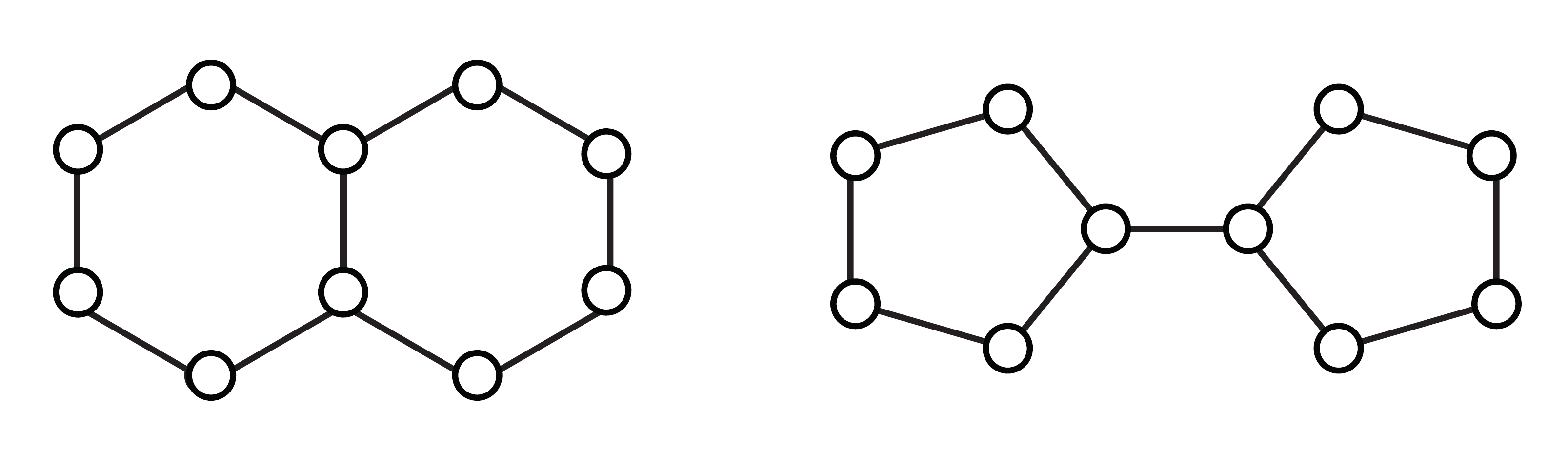}
		\end{center}
		\caption{First pair of the original graphs. 1-WL indistinguishable but not cospectral. \label{fig:1WLindistinguishable}}
	\end{figure}
	
	 	\begin{table}[htbp]
		
		\caption{Experimental study results }
		\setstretch{1.5}
		\centering
		\resizebox{\linewidth}{!}{
			\centering
			\begin{tabular}{cccccccccc}
				
				\toprule[1.1pt]
				\label{tab:real_results1}
				
				\textbf{GNN / Test set}&\textbf{1-WL indistinguishable}      &  \textbf{Cospectral}  \\
				
				\midrule
				\texttt{GIN}         &      64±4    &     93±2  &  \\
				
				\texttt{SP-GIN}&      \textbf{99±5}    &   93±4    &  \\
				\hline
				\texttt{GCN}         &     51±4	& 73±16 	&  \\
				
				\texttt{SP-GCN}&      \textbf{98±6} &	\textbf{92±5}\\
				\hline
				\texttt{GAT} &    50±0 &	49±0  &  \\
				
				\texttt{SP-GAT}         &  \textbf{97±11} &	\textbf{77±16} & \\
				\hline
				\texttt{GraphSAGE} &     49±0 &	 49±0 & \\
				
				\texttt{SP-GraphSAGE} &     \textbf{95±12} & \textbf{91±6} \\
				\bottomrule[1.1pt]
				
			\end{tabular}%
		}
	\end{table}
	
\begin{figure}
		\begin{center}
			\includegraphics[width=\linewidth]{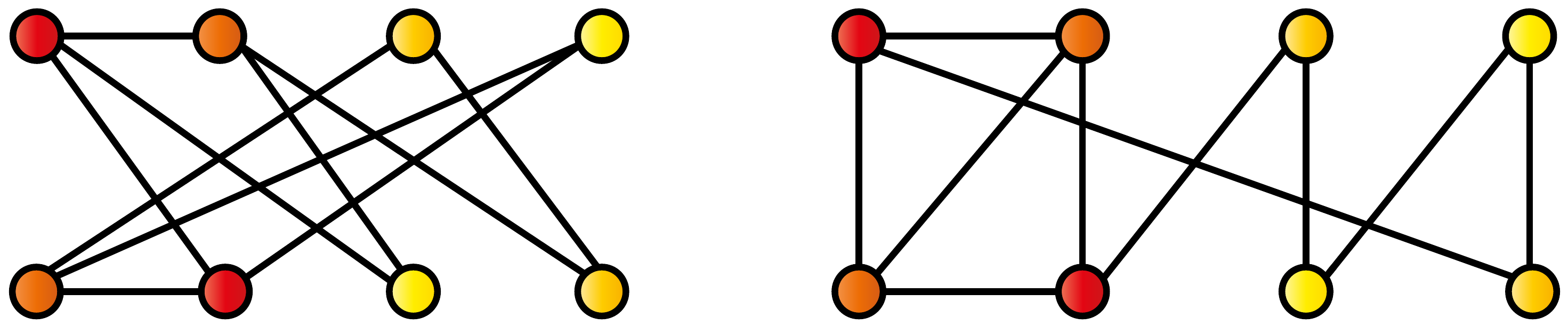}
		\end{center}
		\caption{Cospectral graph coloring based only on the diagonal of the heat kernel.}
			\label{fig:cospectral_coloring_diag}
	\end{figure}

\begin{figure}

		\begin{center}
			\includegraphics[width=\linewidth]{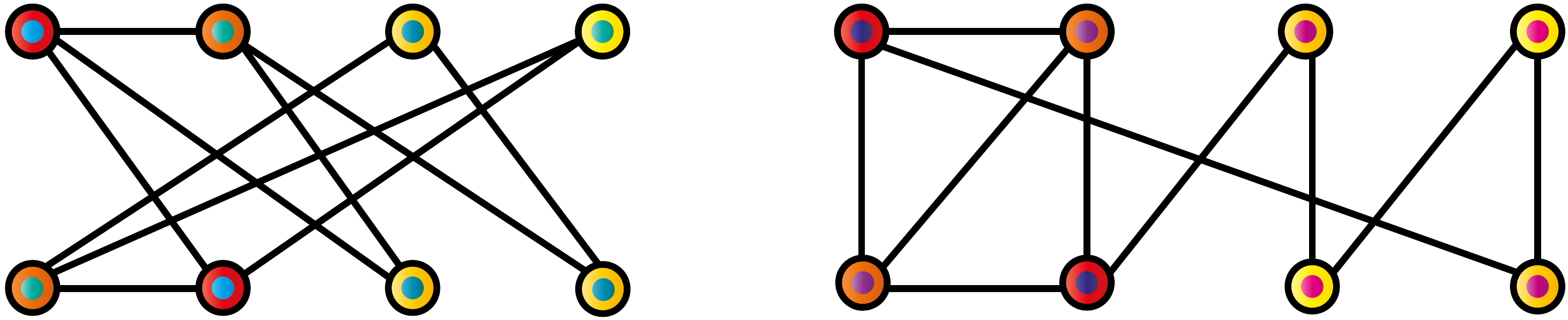}
		\end{center}
		\caption{Cospectral graph coloring based on the diagonal of the heat kernel and the maximum quantile.}
			\label{fig:cospectral_coloring_diag_and_max}
	\end{figure}

	As expected, for the 1-WL indistinguishable pair of graphs, the  MGNNs struggle to identify the source of each graph, because 1-WL cannot differentiate between the sources. The spectral features help them to overcome this issue easily. GIN, which has the most expressive aggregation operation among all the MPGNNs, achieves great accuracy on the cospectral graphs; the other MPGNNs, however, do not. These results make sense, since  cospectral graphs have common structural properties.	In \autoref{fig:cospectral_coloring_diag} and \autoref{fig:cospectral_coloring_diag_and_max} we can see the spectral coloring of the cospectral graphs introduced by the spectral pre-prossessing -- nodes with the same color have the same spectral features. In \autoref{fig:cospectral_coloring_diag} the pre-processing does not use any quantiles and in \autoref{fig:cospectral_coloring_diag_and_max} the pre-processing uses only the maximum quantile. We can see that not only do both colorings \textit{strictly refine} the constant coloring, but that the coloring that uses the maximum quantile \textit{strictly refines} the one that does not.
	\section{Evaluation on real benchmarks}
	We evaluate our pre-processing method on two graph learning tasks: graph classification and node classification. For each task we used four types of GNNs (GIN, GCN, GraphSAGE and GAT) from the Pytorch Geometric framework \cite{fey2019fast} to compare the standard use of the network to our SP method. 
	\subsection{Graph classification}
	 We used nine graph classification benchmarks for this task: five social network datasets (COLLAB, IMDB-BINARY, IMDB-MULTI, REDDITBINARY and REDDIT-MULTI5K), three molecule datasets (MUTAG, PTC, NCI1) and a dataset from the field of bioinformatics (PROTEINS) \cite{Yanardag2015DeepGK}. The task of the benchmarks here is to achieve the highest average validation accuracy with 10-fold cross-validation. We used GNNs with five layers where in each layer's MLP a single hidden layer was used. We used concatenation to create the final graph descriptor and a linear layer to create the final output.  We fine-tuned the dropout of the linear layer to be one of \{0,0.5\}. For the bioinformatics and molecule datasets, we fine-tuned the hidden dimension of all the MLPs to be one of \{16,32\}, while for the social network benchmarks we consistently used a hidden dimension of size 64. The number of epochs that achieved the best cross-validation accuracy, averaged over the 10 folds, was selected. We examined 700 epochs for each configuration. For the SP-MPGNNs we chose the best out of the following two:
	 1. Sampling 10 points in time and not using quantiles at all; 2. Sampling 5 points in time and using the maximum quantile.
	 We report the average validation accuracy and standard deviation over 10 folds.

	 	\begin{table}[htbp]
		
		\caption{Graph classification results -- Molecules and bioinformatics }
		\setstretch{1.5}
		\centering
		\resizebox{\linewidth}{!}{
			\centering
			\begin{tabular}{cccccccccc}
				
				\toprule[1.1pt]
				\label{tab:real_results2}
				
				\textbf{Method}      &  \textbf{MUTAG} &  \textbf{PTC} &  \textbf{PROTEINS} &  \textbf{NCI1} & \\
				
				\midrule
				\texttt{GIN}         &      88±7    &     66±8  &     75±3  &     82±1   &  \\
				
				\texttt{SP-GIN}&      91±6    &   66±7   &   76±3  &     82±1   &  \\
				\hline
				\texttt{GCN}         &      83±6    &	67±6	&   75±3  &     82±1    &  \\
				
				\texttt{SP-GCN}&      \textbf{91±6}    &        68±8 &	75±3 &	81±1&  \\
				\hline
				\texttt{GAT} &     80±9 &	66±9 &	75±3 &	81±1 &  \\
				
				\texttt{SP-GAT}         &  \textbf{90±5} &	68±6 &	75±4 &	81±1 &  \\
				\hline
				\texttt{GraphSAGE} &      83±8 &	65±7 &	73±4 &	82±1 & \\
				
				\texttt{SP-GraphSAGE} &    \textbf{91±7} &	65±6	& 73±4	& 82±1 & \\
				\bottomrule[1.1pt]
				
			\end{tabular}%
		}
	\end{table}
	
	\begin{table}[htbp]
		
		\caption{Graph classification results -- Social networks}
		\setstretch{1.5}
		\centering
		\resizebox{\linewidth}{!}{
			\centering
			\begin{tabular}{cccccccccc}
				
				\toprule[1.1pt]
				\label{tab:real_results3}
				
				\textbf{Method}      & \textbf{COLLAB} & \textbf{IMDB-B} &  \textbf{IMDB-M} &        \textbf{REDDIT-B} &  \textbf{REDDIT-M} \\
				
				\midrule
				\texttt{GIN} &   70±1&	73±3&	50±3&	78±2&	54±1 & \\
				
				\texttt{SP-GIN}         &   \textbf{77±1}&	73±4&	51±4&	\textbf{86±2}&	\textbf{57±2}& \\
				\hline
				\texttt{GCN} &   76±1&	65±3&	41±3&	90±1&	55±1& \\
				
				\texttt{SP-GCN}         &    77±2&	\textbf{74±4}&	\textbf{50±4}&	91±1&	56±1 &\\
				\hline
			\texttt{GAT} &  42±10&	52±3&	36±2&	71±4&	32±5&  \\
				
				\texttt{SP-GAT}         &  \textbf{74±2}&	\textbf{73±4}&	\textbf{50±4}&	\textbf{91±2}&	\textbf{56±1} & \\
				\hline
					\texttt{GraphSAGE} &   40±9&	52±3&	36±2&	73±3&	35±2 &\\
				
				\texttt{SP-GraphSAGE}         &   \textbf{77±2}&	 \textbf{73±3}&	\textbf{50±4}&	\textbf{91±1}&	\textbf{57±2}&\\
				\bottomrule[1.1pt]
				
			\end{tabular}%
		}
	\end{table}
	 In general, the SP-MPGNNs performed better than the MPGNNs, especially on the social network benchmarks that contains no initial features for the nodes.
	\subsection{Node classification}
	We used four node classification benchmarks for this task:  three citation network datasets (Cora, CiteSeer and PubMed) \cite{yang2016revisiting} and a biochemistry dataset (PPI) \cite{zitnik2017predicting}. The task of the benchmarks here is to achieve the highest average test accuracy upon 100 random initializations of the GNNs. For the citation networks, only the number of message passing layers, the hidden dimension of the MLPs and the number of training epochs, were fine-tuned, using the validation set.
	The number of the layers was one of \{2, 3, 4\}, the hidden dimension was one of \{128, 256, 384, 512\} and each model was trained for at most 200 epochs. Specifically for PPI, there were two layers, the hidden dimension was 512 and the models were trained for 800 epochs. The spectral pre-process was calibrated exactly as in the graph classification evaluation.
	We repeated each training-testing session 100 times and report the average accuracy and standard deviation of the test set.
	\begin{table}[htbp]
		
		\caption{Node classification results}
		\setstretch{1.5}
		\centering
		\resizebox{\linewidth}{!}{
			\centering
			\begin{tabular}{cccccccc}
				
				\toprule[1.1pt]
				\label{tab:real_results4}
				
				\textbf{Method}      & \textbf{CiteSeer} & \textbf{Cora} &  \textbf{PubMed} &        \textbf{PPI}  \\
				
				\midrule
				\texttt{GIN} &  	71.9±0.6&		81.8±0.5&		79.6±0.5& 91.1±0.2 & \\
				
				\texttt{SP-GIN}         &   	71.3±0.6&	81.9±1.8& 78.8±0.7 & 91.4±0.2 & \\
				\hline
				\texttt{GCN} &   	63.5±4.4&		78.1±2.6&		80.4±0.5&  88.8±0.1\\
				
				\texttt{SP-GCN}         &    	\textbf{72.1±0.8}&		\textbf{82.3±1.4}&		80.8±0.4&  \textbf{89.2±0.1}& \\
				\hline
				\texttt{GAT} &  	64.1±4.5&		81.6±1.0&	79.9±1.3& 79.6±0.2& \\
				
				\texttt{SP-GAT}         & 	\textbf{72.3±1.5}&	79.2±1.9&		80.4±0.7 & 	\textbf{80.7±0.3}& \\
				\hline
				\texttt{GraphSAGE} &  72.8±0.6&	82.9±0.9&		80.2±0.6& 95.8±0.1 
& \\
				
				\texttt{SP-GraphSAGE}       &	72.9±0.6&		81.9±2.3&		80.8±0.5& \textbf{96.0±0.1}& \\
				\bottomrule[1.1pt]
				
			\end{tabular}%
		}
	\end{table}
	
	 Even though each node in the benchmark contains a feature vector with hundreds of entries, appending to it a relatively small number of spectral features usually improved the accuracy of the MPGNNs. This can be explained by the fact that the spectral features also contain global information about the graph and the node's position according to it. This information cannot be learned using a small amount of message passing iterations. 
	\subsection{Ablation study}
    Alongside the benchmark testing we conducted, we also wanted to examine the effect of the parameters of our spectral pre-processing method, including the number of points in time to sample, the range of the sample and the quantiles to use. For this experiment we chose one benchmark with initial features (NCI1) and one benchmark with none (COLLAB). We again used the four MPGNNs. Each was trained using a hidden dimension of 64 with five message passing layers, for 700 epochs. We report the average and standard deviation of the test accuracy on 10 different folds of the datasets. We report only the best configuration per total size of node features. The configuration setting is reported as `(start of the sampling range in powers of 10, end of the sampling range in powers of 10, number of samples, used quantiles)`.`MMM` in the quantiles entry denotes the use of the max, min and median quantiles. The full results can be found in \autoref{appendix:ablation}.
	
	\begin{table}[htbp]
		
		\caption{Ablation study on the COLLAB results}
		\setstretch{1.5}
		\centering
		\resizebox{\linewidth}{!}{
			\centering
			\begin{tabular}{cccccccc}
				
				\toprule[1.1pt]
				\label{tab:ablation:COLLAB}
				
				\textbf{GNN}  & \textbf{Feature size} & \textbf{Configuration} & \textbf{Accuracy}\\
				
				\midrule
				
				\texttt{SP-GIN}   &   5&	(-1,1,5,none)&0.777±0.018  &  \\
			
				\texttt{SP-GIN}   &   10&	(-1,1,10,none)&0.780±0.021  &  \\
				
				\texttt{SP-GIN}   &   20&	(-1,1,20,max)&0.788±0.021	  & \\
			
				\texttt{SP-GIN}   &   40&	(-2,2,20,max)&0.781±0.020 &  \\
				\hline
				\texttt{SP-GCN}   &   5&	(-2,2,5,none)&0.774±0.025   &  \\
			
				\texttt{SP-GCN}   &   10&	(-2,2,10,none)&0.778±0.019   &  \\
			
				\texttt{SP-GCN}   &   20&	(-2,2,5,MMM)&0.787±0.021  &  \\
			
				\texttt{SP-GCN}   &   40&	(-2,2,10,MMM)&0.786±0.020  &  \\
				\hline
			    \texttt{SP-GAT}   &   5&		(-1,1,5,none)&0.758±0.022  &  \\
			
				\texttt{SP-GAT}   &   10&		(-1,1,10,none)&0.758±0.015   &  \\

				\texttt{SP-GAT}   &   20&	(-2,2,20,none)&0.764±0.016  &  \\
			
				\texttt{SP-GAT}   &   40&		(-3,3,20,max)&0.756±0.015  &  \\
				\hline
				\texttt{SP-GraphSAGE}   &   5&	(-1,1,5,none)&0.787±0.017  &  \\
				
				\texttt{SP-GraphSAGE}   &   10&	(-1,1,10,none)&0.795±0.014  &  \\
				
				\texttt{SP-GraphSAGE}   &   20&	(-3,3,10,max)&0.779±0.015  &  \\
				
				\texttt{SP-GraphSAGE}   &   40&	(-3,3,20,max)&0.780±0.019  &  \\

				\bottomrule[1.1pt]
				
			\end{tabular}%
		}
	\end{table}
		\begin{table}[htbp]
		
		\caption{Ablation study on the NCI1 results}
		\setstretch{1.5}
		\centering
		\resizebox{\linewidth}{!}{
			\centering
			\begin{tabular}{cccccccc}
				
				\toprule[1.1pt]
				\label{tab:ablation:NCI1}
				
				\textbf{GNN}  & \textbf{Features size} & \textbf{Configuration} & \textbf{Accuracy}\\
				
				\midrule
				
				\texttt{SP-GIN}   &   5&	(-1,1,5,none)&0.806±0.024  &  \\
			
				\texttt{SP-GIN}   &   10&	(-2,2,5,max)&0.812±0.014  &  \\
				
				\texttt{SP-GIN}   &   20&	(-2,2,10,max)&0.812±0.014  & \\
			
				\texttt{SP-GIN}   &   40&	(-3,3,20,max)&0.816±0.015  &  \\
				\hline
				\texttt{SP-GCN}   &   5&	(-1,1,5,none)&0.801±0.021  &  \\
			
				\texttt{SP-GCN}   &   10&	(-2,2,5,max)&0.807±0.015  &  \\
			
				\texttt{SP-GCN}   &   20&	(-2,2,20,none)&0.811±0.014&  \\
			
				\texttt{SP-GCN}   &   40&	(-3,3,20,max)&0.804±0.021  &  \\
				\hline
				\texttt{SP-GAT}   &   5&	(-2,2,5,none)&0.806±0.015  &  \\
			
				\texttt{SP-GAT}   &   10&	(-2,2,10,none)&0.807±0.014 &  \\

				\texttt{SP-GAT}   &   20&	(-3,3,10,max)&0.805±0.014  &  \\
			
				\texttt{SP-GAT}   &   40&	(-2,2,20,max)&0.801±0.012  &  \\
				\hline
                \texttt{SP-GraphSAGE}   &   5&	(-1,1,5,none)&0.821±0.010  &  \\
				
				\texttt{SP-GraphSAGE}   &   10&	(-2,2,5,max)&0.816±0.010  &  \\
				
				\texttt{SP-GraphSAGE}   &   20&	(-1,1,10,max)&0.817±0.018  &  \\
				
				\texttt{SP-GraphSAGE}   &   40&	(-2,2,10,MMM)&0.812±0.016   &  \\
				\bottomrule[1.1pt]
				
			\end{tabular}%
		}
	\end{table}
    
    For the NCI1 benchmark we can see that the range and the quantile amount with which we chose to train the MPGNNs in the graph and node classification tasks achieve the best results where there are 10 features. For the NCI1 and COLLAB benchmarks, we can see that the performance of SP-MPGNNs can be improved even further by choosing spectral features with more than 10 entries.

	\section{Related works} \label{sec:related_works}
	This section surveys works related to our research. We split the section into paragraphs by the method used to improve the expressive power of MPGNNs -- spectral methods and methods for generalizing the message passing scheme.  
	\paragraph{Use of spectral decomposition in GNNs.}
	Work has been done to improve the expressive power of GNNs, with some studies adopting the spectral based approaches. An example OF such an approach is the SAN architecture \cite{kreuzer2021rethinking}. First, SAN  finds the spectral decomposition of the graph Laplacian using the kth smallest eigenvalues and their appropriate
    eigenvectors. Then it encodes them into node features using a transformer
    with self-attention. Another example is the DGN architecture \cite{beani2021directional}, which uses the eigendecomposition of the Laplacian to calculate the derivative or direction between the nodes. The directions are then encoded as node features for the GNN. Both SAN and DGN have been shown to produce SOTA results on real world benchmarks. These methods, however, allow different descriptors for isomorphic graphs since they
    are dependent on the eigendecomposition representation.
	\paragraph{WL go X.}
	Several other works tried to break the limit of expressivity of MPGNNs. Some introduce new sophisticated GNN architectures that are based on interesting concepts from various fields of research. These concepts usually generalize the GNNs' message passing scheme, which makes the new architectures more expressive but less efficient.
    The first attempt to extend the expressive power of GNNs was the K-Dimensional Graph Neural Network \cite{morris2019weisfeiler}. These networks generalize MPGNNs in the same way the k-WL test generalizes the WL test. Hence, their expressive power is naturally better, upper bounded by the k-WL instead of the WL. Unfortunately, these networks require tremendous memory and computation time as K increases, similar to the k-WL test. 
    
    
    The Simplicial Isomorphism Network (SIN) \cite{bodnar2021weisfeiler-top} is another approach that extends the expressive power of GNNs. This method treats graphs as a general algebraic object called a simplicial complex and performs the message passing between every two neighbors in the simplicial complex instead of the adjacent vertices. Bodnar et al. proved that SIN is strictly more powerful than the WL test and at least as powerful as the 3-WL test.
    The Cell Isomorphism Network (CIN) \cite{bodnar2021weisfeiler} is yet another architecture based on an algebraic object. This object is called a regular cell complex and it generalizes the simplicial complex. In their paper, the researcher use the definition of `cell complex adjacencies` to define the new scheme of message passing. Similar to SIN, it was proved that CIN is strictly more powerful than the WL test and at least as powerful as the 3-WL test. They also present great results on learning tasks for molecular problems.
	
	\section{Discussion}
	In this work we demonstrated how one can strictly improve the expressive power of the WL test an infinite number of times in the WL hierarchy using the diagonal coloring of the k-WL algorithm, and simultaneously improve the upper bound for MPGNNs, without any change in their architecture.
	We also proposed spectral pre-processing for MPGNNs that is based on the diagonal and quantiles of the heat kernel matrix. From the results of the graph classification and node classification benchmarks, we conclude that our method of pre-processing improves the performance of MPGNNs on real world graph-structured data. For example, the classification accuracy of GIN, the most expressive MPGNN, on social networks graphs, improved by 3.5\% and, when using GAT the improvement is much more significant and stands at 22\%.
	
	In light of our results, we encourage future research of the possible equivariant and insightful pre-coloring or pre-processing that can be done before the learning phase of MPGNNs, similar to the spectral pre-processing we presented.

	\section*{Acknowledgements}
		This research was partially supported by the Technion Hiroshi Fujiwara Cyber Security
        Research Center and the Israel National Cyber Directorate.
	\clearpage
	
	\bibliography{spectral_GIN.bib}
	\bibliographystyle{plainnat}

	\clearpage

	\appendix{}

\renewcommand\thefigure{\thesection.\arabic{figure}} 
\renewcommand\thetable{\thesection.\arabic{table}} 
\renewcommand\theequation{\thesection.\arabic{equation}}  
\setcounter{figure}{0}  
\setcounter{table}{0}

\crefalias{section}{appsec}
\crefalias{subsection}{appsec}
\crefalias{subsubsection}{appsec}

	\section{Proofs}
    In the following proofs we assume the definition of the k-WL as defined in \cite{morris2019weisfeiler}.
    $C^0_{k-WL}$ is defined to be equal between any two tuples of vertices from $\Ga$ and $\Gb$, if and only if the two subgraphs of $\Ga$ and $\Gb$ comprising all the vertcies in each tuple are isomorphic.
    We first define  the multiset for iteration $i$ at index $j$ to be $c^{i,j}_{k-WL}(v_1,...v_k)=\{\{C^{i-1}_{k-WL}(v_1,...v_{j-1},w,v_{j+1},...v_k)| w \in V\}\}$. Finally, we define the k-WL coloring at iteration $i$ on tuple $s$ to be $C^i_{k-WL}(s)=(c^{i,1}_{k-WL}(s),...c^{i,K}_{k-WL}(s))$.
	\subsection{Theorem 1 proof}
	\begin{proof}
	\label{proof:t1}
		\begin{enumerate}[wide, labelwidth=!, labelindent=0pt]
		\item Let $\Ga$ and $\Gb$ be two isomorphic graphs where $\sigma: V_1 \rightarrow V_2$ is the isomorphism.
		We will prove by induction that after $n$ message passing iterations of 1-WL initilized with permutation equivariant coloring, $R_2$, the coloring of every pair $v\in V_1$ and $u\in V_2$ s.t. $\sigma (v)=u$ is the same.
		
		\textbf{Base (n=0):} $R_2$ is permutation equivarinat and hence by its definition  $R_2(v)=R_2(u)$ for each $v\in V_1$ and $u\in V_2$ s.t. $\sigma (v)=u$.
		
		\textbf{Step:} 
		From the induction assumption we know that every two nodes $v\in V_1$ and $u=\sigma (v)\in V_2$ have the same color after $n$ message passing iterations of 1-WL. For each such $v$ and $u$ we will look at the coloring after the $n+1$ iteration of 1-WL. These are equal to  $(C^{n}_{1-R_2WL}(v),\{\{C^{n}_{1-R_2WL}(x) | x \in \N(v)\}\})$ and $(C^{n}_{1-R_2WL}(u),\{\{C^{n}_{1-R_2WL}(x) | x\in\N(u)\}\})$, respectively. $C^{n}_{1-R_2WL}(u)$ and $C^{n}_{1-R_2WL}(v)$ are equal from the induction assumption. $\sigma$ is an isomorphism and hence $x \in \N(v)\iff\sigma(x) \in N(u) $ and, therefore, $\{\{C^{n}_{1-R_2WL}(x) | x \in \N(v)\}\}$ and $\{\{C^{n}_{1-R_2WL}(x) | x \in \N(u)\}\}$ are equal.
		Since $\sigma$ is a bijection, we get that the coloring histogram of $\Ga$ and $\Gb$ is the same for each $n$.
		
		\item Let $\Ga$ and $\Gb$ be any two graphs and let $R_1,R_2$ be two initial colorings for 1-WL s.t. $R_2$ \textit{refines} $R_1$. We will prove by induction that for each $v\in V_1$ and $u\in V_2$, s.t.  $C^{n}_{1-R_2 WL}(v)=C^{n}_{1-R_2WL}(u)$, $u,v$ also satisfy $C^{n}_{1-R_1WL}(v)=C^{n}_{1-R_1WL}(u)$
		for any number $n$ of 1-WL message passing iterations. Therefore, if $C^{n}_{1-R_1WL}(v)\neq C^{n}_{1-R_1WL}(u)$ then $C^{n}_{1-R_2WL}(v)\neq C^{n}_{1-R_2WL}(u)$.
		
		\textbf{Base (n=0):} For any $v\in V_1$ and $u\in V_2$, if $C^{0}_{1-R_2WL}(v)=C^{0}_{1-R_2WL}(u)$ then $C^{0}_{1-R_1WL}(v)=C^{0}_{1-R_1WL}(u)$ since $R_2$ \textit{refines} $R_1$.
		
		\textbf{Step:} Let $v\in V_1$ and $u\in V_2$ be any two vertices s.t. $C^{n+1}_{1-R_2WL}(v)=C^{n+1}_{1-R_2WL}(u)$. Their coloring in the $n+1$ iteration is equal to $(C^{n}_{1-R_2WL}(v),\{\{C^{n}_{1-R_2WL}(x) | x \in \N(v)\}\})$
		and $(C^{n}_{1-R_2WL}(u),\{\{C^{n}_{1-R_2WL}(x) | x \in \N(u)\}\})$, respectively. From the induction assumption we find that $C^{n}_{1-R_1WL}(v)=C^{n}_{1-R_1WL}(u)$. In addition, we know that the two multisets in the second part of the tuples are equal, this means that there exists an injective mapping $\mu:\N(u) \rightarrow \N(v)$ s.t. $C^{n}_{1-R_2WL}(x) = C^{n}_{1-R_2WL}(\mu(x))$ and hence by the induction assumption $\{\{C^{n}_{1-R_1WL}(x) | x \in \N(v)\}\} = \{\{C^{n}_{1-R_1WL}(x) | x \in \N(u)\}\}$ and therefore $C^{n+1}_{1-R_1WL}(v)=C^{n+1}_{1-R_1WL}(u)$.
		
		If $\Ga$ and $\Gb$ are 1-$R_1$WL distinguishable they have different 1-$R_1$WL histograms after some iteration $n$. Hence, there does not exist an injective mapping $\mu:V_1 \rightarrow V_2$ s.t. $C_{1-R_1WL}(x) = C_{1-R_1WL}(\mu(x))$ for any $x\in V_1$. From the claim proved by induction there does not exist an injective mapping $\mu:V_1 \rightarrow V_2$ s.t. $C_{1-R_2WL}(x) = C_{1-R_2WL}(\mu(x))$ for any $x\in V_1$. Therefore $\Ga$ and $\Gb$  have different $1-R_2$WL histograms and are distinguishable by 1-$R_2$WL.
	\end{enumerate}
	\end{proof}
	
	\subsection{Theorem 2 proof}
	\begin{proof}
	\label{proof:l1}
	   Given $\Ga$ and $\Gb$ s.t. $\{\{\Delta$(3-WL)$(v) | v \in V_1\}\}=\{\{\Delta$(3-WL)$(v) | v \in V_2\}\}$ we will prove that  $\{\{C_{3-WL}(x,y,z) | x,y,z \in V_1\}\}=\{\{C_{3-WL}(x,y,z) | x,y,z \in V_2\}\}$. 
	   
	   From the equality of the diagonal colorings histogram we know that there is an injective mapping $\mu:V_1 \rightarrow V_2$ s.t. for any $v \in V_1$, $\Delta$(3-WL)$(v) = \Delta$(3-WL)$(\mu(v))$. From structure of $C_{3-WL}$ we know that $\{\{C_{3-WL}^{n-1}(v,v,z) | z \in V_1 \}\} = \{\{C_{3-WL}^{n-1}(\mu(v),\mu(v),z) | z\in V_2 \}\}$ for any $v\in V_1$. Hence, there is an injective mapping $\mu_2 : V_1\cross V_1 \rightarrow V_2 \cross V_2$  s.t. for any $u,v \in V_1$,
	   $C_{3-WL}^{n-1}(v,v,u) = C_{3-WL}^{n-1}(\mu_2(v,u)_1,\mu_2(v,u))$, and again from the structure of $C_{3-WL}$, the following exists  $\{\{C_{3-WL}^{n-2}(v,y,u) | y\in V_1 \}\} = \{\{C_{3-WL}^{n-2}(\mu_2(v,u)_1,y,\mu_2(v,u)_2) | y\in V_2 \}\}$ for any $u,v \in V_1$. Hence $\{\{C_{3-WL}(x,y,z) | x,y,z \in V_1\}\}=\{\{C_{3-WL}(x,y,z) | x,y,z \in V_2\}\}$, since the 3-WL algorithm converges and we assume it converges after n-2 iterations.
	   The proof can be generalized easily to any K.
	   
	   Given $\Ga$ and $\Gb$ s.t.
	   $\{\{C_{k-WL}(v_1,...v_k) | v_1,...v_k \in V_1\}\}=\{\{C_{k-WL}(v_1,...v_k) | v_1,...v_k \in V_2\}\}$,
	   we will prove that  $\{\{\Delta$(k-WL)$(v) | v \in V_1\}\}=\{\{\Delta$(k-WL)$(v) | v \in V_2\}\}$.
	   From the initialization of k-WL we know that the color of each tuple of the form $(v,..,v)$ is equal only to other tuples of this form since they are the only ones that represents a graph with a single vertex.
	   Hence, for any $v\in V_1$ and $u_1,u_2,...u_k \in V_2$ if   $C_{k-WL}(v,...,v)=C_{k-WL}(u_1,...,u_k)$; then necessarily $u_1=u_2=...=u_k$. Since any $v \in V_1$ is injectively mapped to $u \in V_2$ with the same diagonal coloring, we get that $\{\{\Delta$(k-WL)$(v) | v \in V_1\}\}=\{\{\Delta$(k-WL)$(v) | v \in V_2\}\}$.
	\end{proof}
	
	\subsection{Theorem 3 proof}
	From \autoref{theorm:1} it immediately is derived that 1-$\Delta$(k+1-WL)WL is as expressive at least as 1-$\Delta$(k-WL)WL. To show that this inequality is strict, we will find a pair of graphs for each $K\geq2$ s.t. they are indistinguishable by 1-$\Delta$(k-WL)WL but distinguishable by 1-$\Delta$(k+1-WL)WL. For any $K\geq2$ we know there exists $\Ga$ and $\Gb$ s.t. they are distinguishable by k+1-WL and indistinguishable by k-WL. From  \autoref{theorm:2} we know that this pair of graphs is also distinguishable by the 1-$\Delta$(k+1-WL)WL algorithm. We also know from \autoref{theorm:2} that the $\Delta$(k-WL) histograms of the graphs are equal. We will prove that the  1-$\Delta$(k-WL)WL histograms of the graphs are also equal by showing that the message passing iterations of 1-WL does not change the nodes colors except for the marking/representation of the colors, i.e., the message passing iterations of the 1-WL does not add any new information to the coloring.
	After a single iteration of 1-$\Delta$(k-WL)WL, the new coloring of any vertex $v$ is $(\Delta(k-WL)(v),\{\{\Delta(k-WL)(u)| u \in \N(v)\}\})$, i.e., the new information added to the coloring is the coloring histogram of the neighbors. We will show that this information can be derived from $\Delta(k-WL)(v)$ for any $v$. From the initialization of k-WL we can find any color of a tuple $(v,v,...u)$ such that $u \in \N(v)$ since their representing graphs are isomorphic and different from the representing graphs for $ (v,v,...x)$ where $ x \not\in \N(v)$. In this way we can find any color of a tuple $(v,u,...u)$ s.t. $u \in \N(v)$. Again from the initialization of k-WL we can find the color of any $(u,u,...u)$ s.t. $u \in \N(v)$.
	
	Since the coloring of 1-$\Delta$(k-WL)WL does not change in any iteration  and because the coloring histograms are equal from the beginning,  1-$\Delta$(k-WL)WL cannot distinguish between the pair of graphs.
	
	\subsection{Example 1 proof}
	\begin{proof}
		We will prove that $C_{1-WL}^1\equiv D$, i.e., the coloring generated after a single iteration of 1-WL initialized with constant coloring equals $D$. For any vertex v, it is colored with the following coloring: $(C^0_{1-WL}(v),\{\{C^0_{1-WL}(x) | x\in \N(v)\}\})=(CONST,\{\{CONST,CONST,...CONST\}\})$ where the multiset size is equal to the size of $\N(v)$. Hence $C^1_{1-WL}\equiv D$.
		
		\end{proof}
	\subsection{Theorem 4 proof}
	\begin{proof}
	\label{proof:t3}
	    From \autoref{theorm:1} it is immediately derived that \textit{Spectral WL} is as expressive at least as 1-WL since the spectral pre-coloring is permutation equivariant and any coloring \textit{refines} the constant coloring. We will show that there exist two graphs that are indistinguishable by 1-WL but distinguishable by \textit{Spectral WL} and hence \textit{Spectral WL} is strictly more expressive than 1-WL.
	    
	    Let $\Ga$ and $\Gb$ be the graphs representing the Decalin and Bicyclopentyl molecules (\autoref{fig:1WLindistinguishable}). It was previously shown that $\Ga$ and $\Gb$ are not isomorphic but cannot be distinguished by the 1-WL test \cite{sato2020survey} . Their \textit{Spectral WL} histograms using $m=1$ with $t=1$ and $r=0$ after the initialization phase are shown in \autoref{tab:spectral_histograms}. Since these histograms are different, \textit{Spectral WL} will determine that these graphs are not isomorphic.
		
		\begin{table}
			\centering
			\resizebox{\linewidth}{!}{
			\begin{tabular}{ lcccccc }
				\hline
				\multicolumn{7}{c}{Coloring Histogram} \\
				\hline
				\backslashbox{Graph}{Color} &0.1914& 0.1929& 0.2891& 0.291& 0.3078& 0.3098 \\
				\hline
				$\Ga$  & 2 &0 & 4 & 0 & 4 & 0  \\
				\hline
				$\Gb$   & 0 &2 & 0 & 4 & 0 & 4  \\ 
				\hline
			\end{tabular}
			}
			\caption{Coloring histograms after initialization of \textit{Spectral WL}}
			\label{tab:spectral_histograms}
		\end{table}
	    \end{proof}
	    
	\section{Ablation study -- Full results}
	\label{appendix:ablation}
	    
	 \begin{table}[htbp]
		
		\caption{Ablation study on COLLAB results on GIN}
		\setstretch{1.5}
		\centering
		\resizebox{\linewidth}{!}{
			\centering
			\begin{tabular}{cccccccc}
				
				\toprule[1.1pt]
				
				\textbf{GNN}  & \textbf{Feature size} & \textbf{Configuration} & \textbf{Accuracy}\\
				
				\midrule
				
\texttt{SP-GIN}   &   5  &	(-2,2,5,none)   &   0768±0.015  &  \\
\texttt{SP-GIN}   &   5  &	(-1,1,5,none)   &   0.777±±0.018  &  \\
\texttt{SP-GIN}   &   10  &	(-2,2,5,max)   &   0.777±±0.014  &  \\
\texttt{SP-GIN}   &   10  &	(-2,2,10,none)   &   0.776±±0.017  &  \\
\texttt{SP-GIN}   &   10  &	(-1,1,10,none)   &   0.780±±0.021  &  \\
\texttt{SP-GIN}   &   10  &	(-3,3,10,none)   &   0.763±±0.020  &  \\
\texttt{SP-GIN}   &   10  &	(-1,1,5,max)   &   0.778±±0.022  &  \\
\texttt{SP-GIN}   &   20  &	(-2,2,5,`MMM`)   &   0.755±±0.019  &  \\
\texttt{SP-GIN}   &   20  &	(-2,2,10,max)   &   0.786±±0.020  &  \\
\texttt{SP-GIN}   &   20  &	(-1,1,10,max)   &   0.788±±0.021  &  \\
\texttt{SP-GIN}   &   20  &	(-3,3,10,max)   &   0.772±±0.012  &  \\
\texttt{SP-GIN}   &   20  &	(-2,2,20,none)   &   0.780±±0.018  &  \\
\texttt{SP-GIN}   &   20  &	(-3,3,20,none)   &   0.770±±0.015  &  \\
\texttt{SP-GIN}   &   40  &	(-2,2,10,`MMM`)   &   0.763±±0.017  &  \\
\texttt{SP-GIN}   &   40  &	(-2,2,20,max)   &   0.781±±0.020  &  \\
\texttt{SP-GIN}   &   40  &	(-3,3,20,max)   &   0.770±±0.015  &  \\

			\bottomrule[1.1pt]
				
			\end{tabular}%
        }
    \end{table}
	 \begin{table}[htbp]
		
		\caption{Ablation study on COLLAB results on GCN}
		\setstretch{1.5}
		\centering
		\resizebox{\linewidth}{!}{
			\centering
			\begin{tabular}{cccccccc}
				
				\toprule[1.1pt]
				
				\textbf{GNN}  & \textbf{Feature size} & \textbf{Configuration} & \textbf{Accuracy}\\
				
				\midrule
				
\texttt{SP-GCN}   &   5  &	(-2,2,5,none)   &   0.774±±0.025  &  \\
\texttt{SP-GCN}   &   5  &	(-1,1,5,none)   &   0.760±±0.016  &  \\
\texttt{SP-GCN}   &   10  &	(-2,2,5,max)   &   0.777±±0.024  &  \\
\texttt{SP-GCN}   &   10  &	(-2,2,10,none)   &   0.778±±0.019  &  \\
\texttt{SP-GCN}   &   10  &	(-1,1,10,none)   &   0.762±±0.022  &  \\
\texttt{SP-GCN}   &   10  &	(-3,3,10,none)   &   0.777±±0.019  &  \\
\texttt{SP-GCN}   &   10  &	(-1,1,5,max)   &   0.771±±0.021  &  \\
\texttt{SP-GCN}   &   20  &	(-2,2,5,`MMM`)   &   0.787±±0.021  &  \\
\texttt{SP-GCN}   &   20  &	(-2,2,10,max)   &   0.774±±0.020  &  \\
\texttt{SP-GCN}   &   20  &	(-1,1,10,max)   &   0.770±±0.021  &  \\
\texttt{SP-GCN}   &   20  &	(-3,3,10,max)   &   0.784±±0.023  &  \\
\texttt{SP-GCN}   &   20  &	(-2,2,20,none)   &   0.770±±0.023  &  \\
\texttt{SP-GCN}   &   20  &	(-3,3,20,none)   &   0.780±±0.024  &  \\
\texttt{SP-GCN}   &   40  &	(-2,2,10,`MMM`)   &   0.786±±0.020  &  \\
\texttt{SP-GCN}   &   40  &	(-2,2,20,max)   &   0.769±±0.023  &  \\
\texttt{SP-GCN}   &   40  &	(-3,3,20,max)   &   0.781±±0.023  &  \\

			\bottomrule[1.1pt]
				
			\end{tabular}%
		}
	\end{table}
	 \begin{table}[htbp]
		
		\caption{Ablation study on COLLAB results on GAT}
		\setstretch{1.5}
		\centering
		\resizebox{\linewidth}{!}{
			\centering
			\begin{tabular}{cccccccc}
				
				\toprule[1.1pt]
				
				\textbf{GNN}  & \textbf{Feature size} & \textbf{Configuration} & \textbf{Accuracy}\\
				
				\midrule
				
\texttt{SP-GAT}   &   5  &	(-2,2,5,none)   &   0.753±±0.022  &  \\
\texttt{SP-GAT}   &   5  &	(-1,1,5,none)   &   0.758±±0.022  &  \\
\texttt{SP-GAT}   &   10  &	(-2,2,5,max)   &   0.750±±0.024  &  \\
\texttt{SP-GAT}   &   10  &	(-2,2,10,none)   &   0.753±±0.019  &  \\
\texttt{SP-GAT}   &   10  &	(-1,1,10,none)   &   0.758±±0.015  &  \\
\texttt{SP-GAT}   &   10  &	(-3,3,10,none)   &   0.750±±0.020  &  \\
\texttt{SP-GAT}   &   10  &	(-1,1,5,max)   &   0.758±±0.022  &  \\
\texttt{SP-GAT}   &   20  &	(-2,2,5,`MMM`)   &   0.749±±0.021  &  \\
\texttt{SP-GAT}   &   20  &	(-2,2,10,max)   &   0.763±±0.022  &  \\
\texttt{SP-GAT}   &   20  &	(-1,1,10,max)   &   0.759±±0.018  &  \\
\texttt{SP-GAT}   &   20  &	(-3,3,10,max)   &   0.758±±0.020  &  \\
\texttt{SP-GAT}   &   20  &	(-2,2,20,none)   &   0.764±±0.016  &  \\
\texttt{SP-GAT}   &   20  &	(-3,3,20,none)   &   0.758±±0.022  &  \\
\texttt{SP-GAT}   &   40  &	(-2,2,10,`MMM`)   &   0.764±±0.018  &  \\
\texttt{SP-GAT}   &   40  &	(-2,2,20,max)   &   0.754±±0.018  &  \\
\texttt{SP-GAT}   &   40  &	(-3,3,20,max)   &   0.758±±0.022  &  \\

			\bottomrule[1.1pt]
				
			\end{tabular}%
		}
	\end{table}
	 \begin{table}[htbp]
		
		\caption{Ablation study on COLLAB results on GraphSAGE}
		\setstretch{1.5}
		\centering
		\resizebox{\linewidth}{!}{
			\centering
			\begin{tabular}{cccccccc}
				
				\toprule[1.1pt]
				
				\textbf{GNN}  & \textbf{Feature size} & \textbf{Configuration} & \textbf{Accuracy}\\
				
				\midrule
\texttt{SP-GraphSAGE}   &   5  &	(-2,2,5,none)   &   0.779±±0.018  &  \\
\texttt{SP-GraphSAGE}   &   5  &	(-1,1,5,none)   &   0.787±±0.017  &  \\
\texttt{SP-GraphSAGE}   &   10  &	(-2,2,5,max)   &   0.778±±0.020  &  \\
\texttt{SP-GraphSAGE}   &   10  &	(-2,2,10,none)   &   0.779±±0.016  &  \\
\texttt{SP-GraphSAGE}   &   10  &	(-1,1,10,none)   &   0.795±±0.014  &  \\
\texttt{SP-GraphSAGE}   &   10  &	(-3,3,10,none)   &   0.778±±0.024  &  \\
\texttt{SP-GraphSAGE}   &   10  &	(-1,1,5,max)   &   0.792±±0.019  &  \\
\texttt{SP-GraphSAGE}   &   20  &	(-2,2,5,`MMM`)   &   0.776±±0.021  &  \\
\texttt{SP-GraphSAGE}   &   20  &	(-2,2,10,max)   &   0.774±±0.019  &  \\
\texttt{SP-GraphSAGE}   &   20  &	(-1,1,10,max)   &   0.773±±0.025  &  \\
\texttt{SP-GraphSAGE}   &   20  &	(-3,3,10,max)   &   0.779±±0.015  &  \\
\texttt{SP-GraphSAGE}   &   20  &	(-2,2,20,none)   &   0.778±±0.020  &  \\
\texttt{SP-GraphSAGE}   &   20  &	(-3,3,20,none)   &   0.774±±0.022  &  \\
\texttt{SP-GraphSAGE}   &   40  &	(-2,2,10,`MMM`)   &   0.777±±0.019  &  \\
\texttt{SP-GraphSAGE}   &   40  &	(-2,2,20,max)   &   0.777±±0.019  &  \\
\texttt{SP-GraphSAGE}   &   40  &	(-3,3,20,max)   &   0.780±±0.019  &  \\

			\bottomrule[1.1pt]
				
			\end{tabular}%
		}
	\end{table}
	
	 \begin{table}[htbp]
		
		\caption{Ablation study on NCI1 results on GIN}
		\setstretch{1.5}
		\centering
		\resizebox{\linewidth}{!}{
			\centering
			\begin{tabular}{cccccccc}
				
				\toprule[1.1pt]
				
				\textbf{GNN}  & \textbf{Feature size} & \textbf{Configuration} & \textbf{Accuracy}\\
				
				\midrule
\texttt{SP-GIN}   &   5  &	(-2,2,5,none)   &   0.806±±0.024  &  \\
\texttt{SP-GIN}   &   5  &	(-1,1,5,none)   &   0.809±±0.014  &  \\
\texttt{SP-GIN}   &   10  &	(-2,2,5,max)   &   0.812±±0.014  &  \\
\texttt{SP-GIN}   &   10  &	(-2,2,10,none)   &   0.803±±0.019  &  \\
\texttt{SP-GIN}   &   10  &	(-1,1,10,none)   &   0.803±±0.017  &  \\
\texttt{SP-GIN}   &   10  &	(-3,3,10,none)   &   0.808±±0.016  &  \\
\texttt{SP-GIN}   &   10  &	(-1,1,5,max)   &   0.804±±0.015  &  \\
\texttt{SP-GIN}   &   20  &	(-2,2,5,`MMM`)   &   0.809±±0.018  &  \\
\texttt{SP-GIN}   &   20  &	(-2,2,10,max)   &   0.812±±0.014  &  \\
\texttt{SP-GIN}   &   20  &	(-1,1,10,max)   &   0.810±±0.016  &  \\
\texttt{SP-GIN}   &   20  &	(-3,3,10,max)   &   0.807±±0.015  &  \\
\texttt{SP-GIN}   &   20  &	(-2,2,20,none)   &   0.808±±0.021  &  \\
\texttt{SP-GIN}   &   20  &	(-3,3,20,none)   &   0.811±±0.016  &  \\
\texttt{SP-GIN}   &   40  &	(-2,2,10,`MMM`)   &   0.812±±0.014  &  \\
\texttt{SP-GIN}   &   40  &	(-2,2,20,max)   &   0.807±±0.013  &  \\
\texttt{SP-GIN}   &   40  &	(-3,3,20,max)   &   0.816±±0.015  &  \\

			\bottomrule[1.1pt]
				
			\end{tabular}%
		}
	\end{table}
	
	 \begin{table}[htbp]
		
		\caption{Ablation study on NCI1 results on GCN}
		\setstretch{1.5}
		\centering
		\resizebox{\linewidth}{!}{
			\centering
			\begin{tabular}{cccccccc}
				
				\toprule[1.1pt]
				
				\textbf{GNN}  & \textbf{Feature size} & \textbf{Configuration} & \textbf{Accuracy}\\
				
				\midrule
\texttt{SP-GCN}   &   5  &	(-2,2,5,none)   &   0.801±±0.019  &  \\
\texttt{SP-GCN}   &   5  &	(-1,1,5,none)   &   0.801±±0.021  &  \\
\texttt{SP-GCN}   &   10  &	(-2,2,5,max)   &   0.807±±0.015  &  \\
\texttt{SP-GCN}   &   10  &	(-2,2,10,none)   &   0.804±±0.010  &  \\
\texttt{SP-GCN}   &   10  &	(-1,1,10,none)   &   0.803±±0.016  &  \\
\texttt{SP-GCN}   &   10  &	(-3,3,10,none)   &   0.795±±0.016  &  \\
\texttt{SP-GCN}   &   10  &	(-1,1,5,max)   &   0.806±±0.021  &  \\
\texttt{SP-GCN}   &   20  &	(-2,2,5,`MMM`)   &   0.810±±0.014  &  \\
\texttt{SP-GCN}   &   20  &	(-2,2,10,max)   &   0.805±±0.011  &  \\
\texttt{SP-GCN}   &   20  &	(-1,1,10,max)   &   0.803±±0.018  &  \\
\texttt{SP-GCN}   &   20  &	(-3,3,10,max)   &   0.805±±0.017  &  \\
\texttt{SP-GCN}   &   20  &	(-2,2,20,none)   &   0.811±±0.014  &  \\
\texttt{SP-GCN}   &   20  &	(-3,3,20,none)   &   0.810±±0.016  &  \\
\texttt{SP-GCN}   &   40  &	(-2,2,10,`MMM`)   &   0.801±±0.017  &  \\
\texttt{SP-GCN}   &   40  &	(-2,2,20,max)   &   0.797±±0.023  &  \\
\texttt{SP-GCN}   &   40  &	(-3,3,20,max)   &   0.804±±0.021  &  \\

			\bottomrule[1.1pt]
				
			\end{tabular}%
		}
	\end{table}
	
	 \begin{table}[htbp]
		
		\caption{Ablation study on NCI1 results on GAT}
		\setstretch{1.5}
		\centering
		\resizebox{\linewidth}{!}{
			\centering
			\begin{tabular}{cccccccc}
				
				\toprule[1.1pt]
				
				\textbf{GNN}  & \textbf{Feature size} & \textbf{Configuration} & \textbf{Accuracy}\\
				
				\midrule
\texttt{SP-GAT}   &   5  &	(-2,2,5,none)   &   0.806±±0.015  &  \\
\texttt{SP-GAT}   &   5  &	(-1,1,5,none)   &   0.801±±0.018  &  \\
\texttt{SP-GAT}   &   10  &	(-2,2,5,max)   &   0.806±±0.017  &  \\
\texttt{SP-GAT}   &   10  &	(-2,2,10,none)   &   0.807±±0.014  &  \\
\texttt{SP-GAT}   &   10  &	(-1,1,10,none)   &   0.790±±0.016  &  \\
\texttt{SP-GAT}   &   10  &	(-3,3,10,none)   &   0.806±±0.021  &  \\
\texttt{SP-GAT}   &   10  &	(-1,1,5,max)   &   0.800±±0.018  &  \\
\texttt{SP-GAT}   &   20  &	(-2,2,5,`MMM`)   &   0.791±±0.018  &  \\
\texttt{SP-GAT}   &   20  &	(-2,2,10,max)   &   0.792±±0.015  &  \\
\texttt{SP-GAT}   &   20  &	(-1,1,10,max)   &   0.794±±0.021  &  \\
\texttt{SP-GAT}   &   20  &	(-3,3,10,max)   &   0.805±±0.014  &  \\
\texttt{SP-GAT}   &   20  &	(-2,2,20,none)   &   0.798±±0.013  &  \\
\texttt{SP-GAT}   &   20  &	(-3,3,20,none)   &   0.798±±0.013  &  \\
\texttt{SP-GAT}   &   40  &	(-2,2,10,`MMM`)   &   0.801±±0.011  &  \\
\texttt{SP-GAT}   &   40  &	(-2,2,20,max)   &   0.801±±0.012  &  \\
\texttt{SP-GAT}   &   40  &	(-3,3,20,max)   &   0.797±±0.021  &  \\

			\bottomrule[1.1pt]
				
			\end{tabular}%
		}
	\end{table}
	
	 \begin{table}[htbp]
		
		\caption{Ablation study on NCI1 results on GraphSAGE}
		\setstretch{1.5}
		\centering
		\resizebox{\linewidth}{!}{
			\centering
			\begin{tabular}{cccccccc}
				
				\toprule[1.1pt]
				
				\textbf{GNN}  & \textbf{Feature size} & \textbf{Configuration} & \textbf{Accuracy}\\
				
				\midrule
\texttt{SP-GraphSAGE}   &   5  &	(-2,2,5,none)   &   0.812±±0.010  &  \\
\texttt{SP-GraphSAGE}   &   5  &	(-1,1,5,none)   &   0.821±±0.011  &  \\
\texttt{SP-GraphSAGE}   &   10  &	(-2,2,5,max)   &   0.816±±0.010  &  \\
\texttt{SP-GraphSAGE}   &   10  &	(-2,2,10,none)   &   0.810±±0.021  &  \\
\texttt{SP-GraphSAGE}   &   10  &	(-1,1,10,none)   &   0.810±±0.010  &  \\
\texttt{SP-GraphSAGE}   &   10  &	(-3,3,10,none)   &   0.811±±0.009  &  \\
\texttt{SP-GraphSAGE}   &   10  &	(-1,1,5,max)   &   0.810±±0.013  &  \\
\texttt{SP-GraphSAGE}   &   20  &	(-2,2,5,`MMM`)   &   0.807±±0.010  &  \\
\texttt{SP-GraphSAGE}   &   20  &	(-2,2,10,max)   &   0.809±±0.011  &  \\
\texttt{SP-GraphSAGE}   &   20  &	(-1,1,10,max)   &   0.817±±0.018  &  \\
\texttt{SP-GraphSAGE}   &   20  &	(-3,3,10,max)   &   0.811±±0.013  &  \\
\texttt{SP-GraphSAGE}   &   20  &	(-2,2,20,none)   &   0.815±±0.018  &  \\
\texttt{SP-GraphSAGE}   &   20  &	(-3,3,20,none)   &   0.816±±0.013  &  \\
\texttt{SP-GraphSAGE}   &   40  &	(-2,2,10,`MMM`)   &   0.812±±0.016  &  \\
\texttt{SP-GraphSAGE}   &   40  &	(-2,2,20,max)   &   0.810±±0.013  &  \\
\texttt{SP-GraphSAGE}   &   40  &	(-3,3,20,max)   &   0.806±±0.014  &  \\

			\bottomrule[1.1pt]
				
			\end{tabular}%
		}
	\end{table}

\end{document}